\documentclass[lettersize,journal]{IEEEtran}
\usepackage{amsmath,amsfonts,amssymb}
\usepackage{array}
\usepackage{booktabs}
\usepackage{cite}
\usepackage{graphicx}
\usepackage{multirow}
\usepackage{textcomp}
\usepackage{url}
\usepackage{xcolor}

\begin{document}

\title{NaviScale: Generating Large-Scale Semantic Map Datasets for Object Navigation}

\author{Chuanlin Lan, Yanwei Zheng, Yuxi Jing, Weijian Liu, Zhitong Zhou,
Jiarui Fan, Fuzhen Zhuang, Xiao Zhang, and Dongxiao Yu%
\thanks{Chuanlin Lan, Yanwei Zheng, Yuxi Jing, Weijian Liu, Zhitong Zhou, Jiarui Fan, and Xiao Zhang are with the School of Computer Science and Technology, Shandong University, Qingdao, China (e-mail: lancl@sdu.edu.cn; zhengyw@sdu.edu.cn; jyx@mail.sdu.edu.cn; lwj12138@mail.sdu.edu.cn; ztzhou@mail.sdu.edu.cn; 202400130005@mail.sdu.edu.cn; xiaozhang@sdu.edu.cn).}%
\thanks{Fuzhen Zhuang is with the Institute of Artificial Intelligence, Beihang University, Beijing, China (e-mail: zhuangfuzhen@buaa.edu.cn).}%
\thanks{Dongxiao Yu is with the School of Cryptologic Science and Engineering, Shandong University, Jinan 250101, China (e-mail: dxyu@sdu.edu.cn).}%
\thanks{Yanwei Zheng is the corresponding author.}}

\markboth{IEEE Transactions on Multimedia}%
{Lan \MakeLowercase{\textit{et al.}}: NaviScale}

\maketitle

\begin{abstract}
Embodied navigation requires spatial representations that generalize across unseen environments, yet collecting large amounts of annotated data from real 3D environments is difficult. We propose NaviScale for semantic-map-based object navigation (ObjectNav), whose predictor can be trained on pairs of partial and complete semantic maps without reconstructing a complete 3D environment for every training sample. The framework generates large-scale semantic map training data by composing floorplans of real homes with room-level semantic and obstacle maps extracted from MP3D and HM3DSem. NaviScale increases data diversity in two ways: inter-room scaling increases floorplan-level structural diversity, while intra-room scaling fills each fixed floorplan with different combinations of room maps matched by room category. Visibility through Ray Casting (VisRC) converts the composed maps into partial observations that account for field of view, sensing range, and occlusion. The resulting dataset contains 192,000 semantic maps generated from 24,000 floorplans associated with 12,794 properties. With 300k training iterations and the training and inference settings described in this paper, the system reaches 64.3\% SR and 34.8\% SPL on HM3D, together with 43.1\% SR and 16.8\% SPL on MP3D, without changing the prediction architecture. Additional experiments evaluate the quality of the composed maps, the effects of semantic-segmentation errors, and deployment on a physical robot.
\end{abstract}


\section{Introduction}

Embodied navigation in unknown environments requires agents to learn spatial regularities that transfer to unseen layouts~\cite{zhang2021hierarchical,du2020learning,dang2023search,dang2022unbiased}. Collecting enough training data to learn these spatial relationships is difficult because collecting and annotating real 3D environments requires physical access, specialized capture equipment, reconstruction, and semantic labeling. This shortage limits the diversity of environments available for learning transferable spatial relationships.

We address the training-data bottleneck in semantic-map-based ObjectNav. Its core prediction module infers target-object locations from an incomplete semantic map accumulated during exploration and is trained using the corresponding complete semantic map. The training data must therefore expose the predictor to spatial relationships that remain useful when only part of an unfamiliar environment has been observed.

\begin{figure}[!t]
  \centering
  \includegraphics[width=\linewidth]{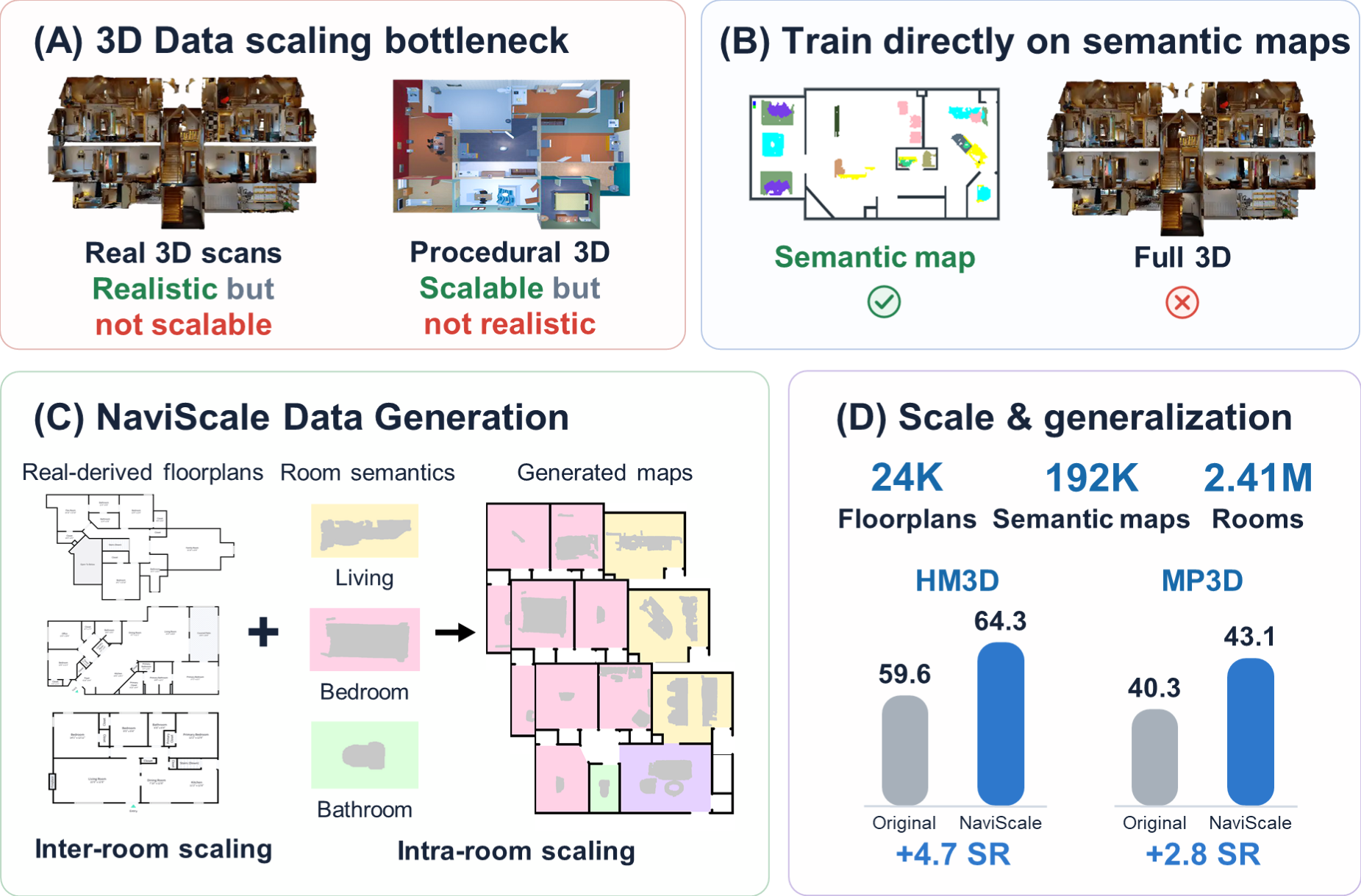}
  \caption{Motivation and overview of NaviScale. (A) Real 3D scans capture real environments but are costly to scale, whereas procedural 3D environments are scalable but depend on predefined generation rules. (B) For semantic-map-based ObjectNav, the spatial predictor can be trained directly from partial and full semantic maps without constructing a complete 3D environment for every training sample. (C) NaviScale combines floorplans of real homes with room-level semantic and obstacle maps with matching room categories to generate large-scale training maps. (D) The resulting dataset contains 24,000 floorplans, 192,000 semantic maps, and 2.41 million room placements. Under the same prediction architecture and the same training and inference settings within each benchmark, replacing the original benchmark training data with NaviScale improves five-run mean SR from 59.6\% to 64.3\% on HM3D and from 40.3\% to 43.1\% on MP3D.}
  \label{fig:first_fig}
\end{figure}

Current semantic map-based navigation methods typically employ deep architectures adapted from semantic segmentation. For instance, NaviFormer~\cite{xie2025naviformer} uses a Transformer with cross-attention to predict long-term goals. In visual recognition, comparable architectures are trained on large annotated datasets such as COCO~\cite{lin2014microsoft}. Yet their navigation counterparts must learn spatial relationships from only hundreds of annotated scene-level semantic maps in MP3D~\cite{chang2017matterport3d} and HM3DSem~\cite{yadav2022habitat}. This contrast motivates expanding navigation training data to better support these models, but the cost of 3D scanning, reconstruction, and semantic annotation makes direct scene-level scaling difficult.

Several approaches address this limitation. ProcTHOR~\cite{deitke2022} creates thousands of synthetic environments using procedural rules. These rules may omit spatial patterns in real buildings, limiting the usefulness of the generated data for learning navigation policies that generalize to real environments. Another line of work~\cite{zhai2023peanut,ramakrishnan2022poni} increases trajectories within existing scenes without introducing new floorplan structures.

These limitations motivate a different way to expand training environments. The spatial predictor requires pairs of partial and complete semantic maps, not a new photorealistic 3D reconstruction for every training scene. We therefore propose NaviScale to generate these map pairs directly. Floorplans of real homes provide room boundaries and their spatial arrangement, while room maps extracted from real 3D scans supply object and obstacle layouts. Matching the room categories connects these two sources: each floorplan room receives the contents of a scanned room of the corresponding type.

NaviScale expands spatial diversity at two levels: overall floorplan structure and object and obstacle layouts within rooms. Accordingly, inter-room scaling introduces more distinct floorplans, while intra-room scaling creates different room-map compositions within each fixed floorplan. The resulting complete maps provide training targets, but the predictor must also learn from what an exploring agent can observe. For each composed map, VisRC (Visibility through Ray Casting) therefore produces partial observations subject to field of view, sensing range, and occlusion.

Because additional maps can come from either new floorplans or new compositions of existing floorplans, our ablations examine how to allocate a fixed data budget. Increasing the number of distinct floorplans is generally more effective than increasing the number of room recompositions per floorplan. A separate partial-map ablation shows that observing complete rooms while leaving other rooms unseen performs better than training primarily from fragmented intra-room views. Together, these results motivate greater emphasis on floorplan-level structural diversity and unseen-room prediction.

Our contributions are threefold: 
\begin{itemize}
\item We generate training data directly as pairs of partial and complete semantic maps for ObjectNav, rather than reconstructing complete 3D environments, producing 192,000 maps from 24,000 floorplans of real homes without changing the prediction architecture.
\item We increase spatial diversity in two ways. Inter-room scaling varies floorplan-level structure, including room count, size, shape, adjacency, connectivity, corridors, and overall layout; intra-room scaling varies object and obstacle layouts drawn from rooms of the same category while holding the floorplan fixed. VisRC converts the composed full maps into visibility-constrained partial observations.
\item Controlled ablations indicate that, when the total number of generated maps is fixed, using more distinct floorplans is more effective than repeated room recomposition or additional trajectories within the same layouts. Partial-map ablations provide complementary evidence that supervision for unseen-room prediction is particularly useful for navigation.

\end{itemize}

\begin{table}[!t]
\caption{Dataset Comparison. Scn., Rm., and Hrs. denote scenes, rooms, and annotation hours; area is in m$^2$. R, S, and RS denote real, simulated, and synthetic data constructed from real sources. ScanNet's 707 environments contain 1,513 scans. For NaviScale, FP reports unique floorplans, base room masks, and unique area; $K=8$ reports generated maps, room placements, and area summed over eight compositions per floorplan. Both rows share the same 1.3k hours of floorplan room-mask processing and labeling, excluding source-dataset annotation. HM3D has no semantic labels.}
\label{tab:dataset}
\setlength{\tabcolsep}{2.1pt}
\centering
\makebox[\linewidth][c]{%
\begin{tabular}{@{}lrrrcc@{}}
\toprule
Data & Scn. & Rm. & Area & Src. & Hrs. \\
\midrule
Replica~\cite{straub2019replica}      & 18   & 35     & 2.19k   & R  & -- \\
MP3D~\cite{chang2017matterport3d}     & 90   & 2,056  & 101.82k & R  & -- \\
ScanNet~\cite{dai2017scannet}         & 707  & --     & 39.98k  & R  & -- \\
RoboTHOR~\cite{deitke2020robothor}    & 75   & --     & 3.17k   & S  & -- \\
ProcTHOR~\cite{deitke2022}            & 12k  & 45.16k & 727k    & S  & -- \\
HM3D~\cite{ramakrishnan2021hm3d}      & 1k   & --     & 365.42k & R  & -- \\
HM3DSem v0.1                            & 120  & --     & --      & R  & -- \\
HM3DSem v0.2~\cite{yadav2022habitat}  & 216  & 3.1k   & --      & R  & 14.2k+ \\
Gibson~\cite{xia2018gibson}           & 572  & --     & 211k    & R  & -- \\
\midrule
Ours (FP)                               & 24k  & 301.66k & 3.38M  & R  & 1.3k \\
Ours ($K=8$)                            & 192k & 2.41M  & 27.05M  & RS & 1.3k \\
\bottomrule
\end{tabular}%
}
\end{table}

\begin{figure*}[!t]
\centering
\includegraphics[width=2\columnwidth]{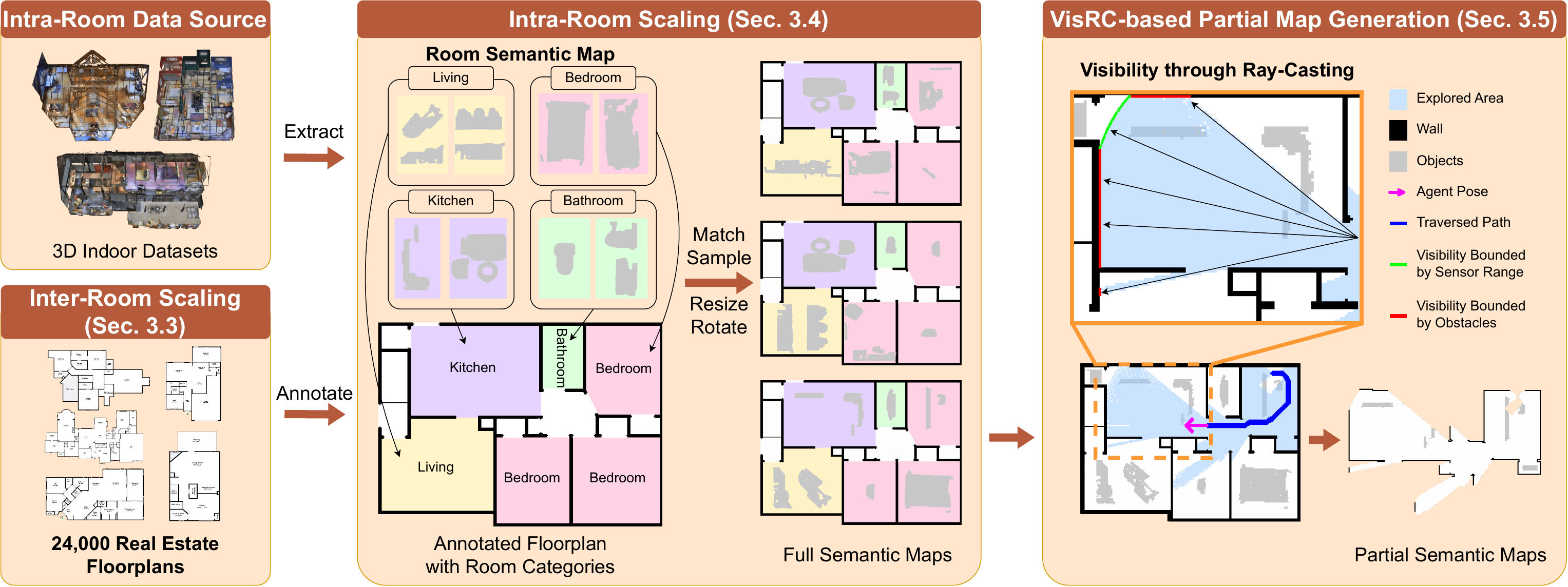}
\caption{Overview of NaviScale. Inter-room scaling collects distinct floorplans of real homes to vary global structure. Intra-room scaling assigns different semantic and obstacle room maps with matching room categories to each fixed layout, generating multiple full maps ($S^{\text{full}}, O^{\text{full}}$). VisRC then applies field-of-view, sensing-range, and occlusion constraints to produce partial observations ($M_t$) for training.}
\label{fig:pipeline}
\end{figure*}

\section{Related Work}
\subsection{Object Navigation Methods}
\label{sec:related_methods}
Semantic map-based methods explicitly maintain a top-down semantic map that accumulates observations over time. At each timestep, the agent receives RGB-D observations, applies semantic segmentation to obtain pixel-wise labels, and projects the labeled pixels into 3D space using depth. The projected points are then accumulated into a 2D top-down grid map, where each cell stores the semantic category of the object at that location. This local observation is merged with the global map from previous timesteps to form an incrementally updated semantic map.

Based on this representation, current methods mainly improve object navigation from two directions: better use of map information and incorporation of external priors.

For map utilization, PONI~\cite{ramakrishnan2022poni} uses UNet~\cite{ronneberger2015u} to score the boundaries of observed partial maps for goal selection, while PEANUT~\cite{zhai2023peanut} adopts PSPNet~\cite{zhao2017pyramid} to directly predict potential target locations. More recently, NaviFormer~\cite{xie2025naviformer} introduces specialized encoders for different semantic map channels and applies cross-attention mechanisms to predict long-term goals.

External priors have mainly been introduced through language models. L3MVN~\cite{yu2023l3mvn} constructs natural language descriptions for map boundaries based on nearby objects and uses a fine-tuned RoBERTa~\cite{liu2019roberta} model to predict which boundary is closer to the target. SGM~\cite{zhang2024imagine} performs multimodal fusion of GPT-4~\cite{achiam2023gpt} or ChatGLM~\cite{glm2024chatglm} outputs with map information. Zero-shot methods such as VoroNav~\cite{wu2024voronav} and TopV-Nav~\cite{zhong2024topv} convert map observations into text and prompt large language models for navigation reasoning.

While language-model priors can improve navigation, they also introduce extra computational cost, and textual descriptions often struggle to preserve the rich spatial structure encoded in semantic maps. This motivates our data-centric approach: instead of relying on external priors, we scale the semantic map training data itself.

\subsection{Datasets for Object Navigation}
\label{sec:related_datasets}
Existing datasets for object navigation can be roughly divided into real-world 3D scanned datasets and synthetic procedural datasets.

Real-world datasets such as Replica~\cite{straub2019replica}, MP3D~\cite{chang2017matterport3d}, HM3D~\cite{ramakrishnan2021hm3d}, ScanNet~\cite{dai2017scannet}, and Gibson~\cite{xia2018gibson} are built by physically scanning indoor environments with specialized capture devices. This pipeline requires access to real buildings, trained operators, and heavy post-processing to reconstruct 3D meshes. Semantic annotation is even more expensive: HM3DSem v0.2~\cite{yadav2022habitat}, for example, required more than 14,200 hours of voxel-level 3D labeling.

Synthetic datasets such as ProcTHOR~\cite{deitke2022} and RoboTHOR~\cite{deitke2020robothor} improve scalability through procedural generation with manually designed rules over room layouts, object placements, and architectural constraints. Although these methods can generate many scenes, hand-designed rules may omit structural and semantic patterns found in real environments, contributing to a domain gap during transfer.

NaviScale takes a different path by combining floorplans of real homes for inter-room layouts with room semantic maps extracted from real 3D scans for intra-room object and obstacle configurations. Because the core predictor operates on 2D spatial semantics rather than complete 3D reconstruction, this design retains overall layouts from real floorplans and local object and obstacle arrangements from scanned rooms while scaling the dataset from hundreds of scenes to 192,000 semantic maps. Additional details on data collection and annotation are provided in the supplementary material.

\section{Method}

\subsection{Problem Formulation}

We formulate semantic map-based object navigation as follows: An agent operates in an environment with discrete actions $\mathcal{A} = \{\text{stop}, \text{forward}, \text{turn left}, \text{turn right}, \text{look up}, \text{look down}\}$. At each timestep $t$, the agent occupies position $(x_t, y_t, o_t)$ where $(x_t, y_t)$ represents planar coordinates and $o_t$ denotes orientation. The environment returns RGB image $I_t^{\text{rgb}}$ and depth image $I_t^{\text{depth}}$. 

The agent maintains a global semantic map $M_t \in \mathbb{R}^{H \times W \times (4+N)}$ that is initialized as an empty map $M_0$ at the beginning of each episode. Following the standard configuration in prior work~\cite{zhai2023peanut,ramakrishnan2022poni}, each pixel in the semantic map represents a 5cm $\times$ 5cm spatial region, providing sufficient resolution for navigation tasks while maintaining computational efficiency. At each timestep $t$, the agent performs semantic segmentation on $I_t^{\text{rgb}}$ to obtain semantic labels $I_t^{\text{semantic}}$ with $N$ semantic channels, then projects these labels into 3D space using $I_t^{\text{depth}}$ to construct a local semantic map $M_t^{\text{local}}$. This local map is merged with the accumulated semantic map $M_{t-1}$ to form the updated semantic map $M_t$. The semantic map $M_t$ contains $4+N$ channels: obstacle map, explored map, current position, history path, and $N$ semantic categories.

We define $M_t$ as the partial map accumulated by the agent up to timestep $t$, representing the incomplete spatial knowledge obtained through exploration. A full map $S^{\text{full}} \in \mathbb{R}^{H \times W \times N}$ represents the complete semantic map containing only the $N$ semantic channels without navigation-specific information. The core prediction task is: given partial map $M_t$, predict the full semantic map using a neural network $f$:
\begin{equation}
S_t^{\text{pred}} = f(M_t)
\label{eq:map_method}
\end{equation}
where $S_t^{\text{pred}}$ represents the predicted semantic map. During training, $S_t^{\text{pred}}$ is supervised to approximate $S^{\text{full}}$. For navigation, the agent extracts target object locations from $S_t^{\text{pred}}$, computes Fast Marching Method (FMM)~\cite{sethian1999fast} distances from current position to target locations accounting for obstacles, and selects the next action accordingly.

\subsection{NaviScale Framework Overview}

As shown in Equation \ref{eq:map_method}, model training requires only pairs of partial and full semantic maps $(M_t, S^{\text{full}})$, without dependency on RGB-D observations that necessitate expensive 3D scanning. This key insight motivates our scalable data generation approach: instead of scaling expensive 3D scene reconstructions, we can directly scale semantic map data.

To construct a complete map, we need both a floorplan that specifies where rooms are and room contents that specify where objects and obstacles are. We obtain the former from floorplans of real homes and the latter from scanned rooms. Room-category matching makes the composition meaningful: a kitchen mask is populated with a kitchen map rather than the contents of an unrelated room type. This separates two choices in the generation process without assuming that room structure and contents are statistically independent in real buildings.

Inter-room Scaling expands the floorplan collection, while Intra-room Scaling selects different room-level semantic and obstacle maps from MP3D and HM3DSem\_v0.1 for each fixed floorplan. The composition yields complete semantic maps $S^{\text{full}}$ and obstacle maps $O^{\text{full}}$. To obtain the corresponding partial maps $M_t$, we sample exploration trajectories and apply VisRC (Visibility through Ray Casting) with field-of-view, range, and occlusion constraints. The following subsections describe these steps in order. The goal is useful navigation supervision rather than exact reproduction of the real-world distribution; we assess that usefulness through navigation in unseen scanned environments.

\subsection{Inter-Room Scaling}
We first build the floorplan collection that determines the global structure of the generated maps. We use inter-room diversity to mean floorplan-level structural diversity, including room count, room size and shape, adjacency and connectivity, corridors, and overall layout. Inter-room scaling expands this diversity by increasing the number of distinct floorplans.

We collected 12,794 unique residential properties from publicly accessible real-estate listings, yielding 24,000 floorplans after separating the floors of multi-story properties. The retained representation contains abstract room geometry and categorical room labels rather than addresses, photographs, personal information, or other identifying attributes. Further collection and ethical details are provided in the supplementary material.

The source SVG files provide room masks, room descriptions, and physical-scale information for metric room dimensions and area calculations. We extract masks from SVG paths and manually assign 12 labels after merging Transition into Others: Bedroom, Living, Dining, Kitchen, Bathroom, Work\&Study, Recreation, Utility, Garage, Stairs\&Hall, Outdoors, and Others. Eleven categories are used for room-map composition; Others regions do not receive a source room map. This processing required about 1,300 human-hours.

After processing, we obtain a floorplan collection $\mathcal{F} = \{F_j\}_{j=1}^{|\mathcal{F}|}$. Each floorplan $F_j \in \mathcal{F}$ contains room boundary masks $\{B_{j,k}\}_{k=1}^{|F_j|}$ and room categories $\{c_{j,k}\}_{k=1}^{|F_j|}$, where $c_{j,k} \in \mathcal{C}$ and $\mathcal{C}$ represents the 11 functional types. Others regions remain unfilled during room-map composition.
The 301,663 base room masks yield 2,413,304 room-map placements at $K=8$. The 24,000 unique floorplans cover 3,381,405~m$^2$; counting the footprint of each of the eight composed maps gives the aggregate 27,051,240~m$^2$ reported in Table~\ref{tab:dataset}. The category-level distribution is provided in the supplementary material.

\subsection{Intra-Room Scaling}
Holding each floorplan and its room boundaries fixed, intra-room scaling varies the room maps with matching room categories assigned to those boundaries. We extract these semantic and obstacle maps from MP3D and HM3DSem\_v0.1, retaining room-level object layouts and obstacle configurations derived from real 3D scans. Room maps used for NaviScale generation come exclusively from the training splits, preventing leakage into the validation scenes.

We extract room-level maps $\mathcal{R} = \{R_{c,i} : c \in \mathcal{C}, i \in [1, N_c]\}$ for each room category $c$, where $N_c$ represents the number of extracted rooms for category $c$. Each room map $R_{c,i} = \{S_{c,i}, O_{c,i}\}$ contains both semantic map $S_{c,i}$ and obstacle map $O_{c,i}$ with object layouts and obstacle configurations extracted from real scanned environments. The source library contains 1,364 MP3D rooms with native room labels and 1,186 valid HM3DSem v0.1 semantic regions with manually assigned room-category labels. The unified 12-category mapping and per-category counts are provided in the supplementary material.

For each floorplan $F_j \in \mathcal{F}$ with room boundaries $\{B_{j,1}, \ldots, B_{j,|F_j|}\}$ and corresponding room categories $\{c_{j,1}, \ldots, c_{j,|F_j|}\}$, we fill each room by selecting room maps from $\mathcal{R}$ that match the room's functional category. 

To construct a complete scene layout for each floorplan $F_j$, we generate multiple combinations of its constituent room maps. The $k$-th combination is
\begin{equation}
\mathbf{s}_{j,k}
= \left(s_{j,k}^{(1)},\ldots,s_{j,k}^{(|F_j|)}\right),
\qquad
s_{j,k}^{(l)} \sim \mathcal{R}_{c_{j,l}},
\label{eq:room_combination}
\end{equation}
where $s_{j,k}^{(l)}$ is the map selected for the $l$-th room and $\mathcal{R}_{c_{j,l}}=\{R_{c_{j,l},i}\}_{i=1}^{N_{c_{j,l}}}$ is the source-map set for its functional category.

By sampling multiple such combinations (up to $K$ per floorplan), we obtain combinations of room maps with matching room categories. We spatially align and scale the selected maps to fit the room boundaries $\{B_{j,1}, \ldots, B_{j,|F_j|}\}$, then verify that the resulting navigable area is sufficiently large and connected. The supplementary material details both procedures. This produces complete scenes $\mathcal{G} = \{G_{j,k} : j \in [1, |\mathcal{F}|], k \in [1, K]\}$. Each $G_{j,k}$ comprises a full semantic map $S^{\text{full}}$ and obstacle map $O^{\text{full}}$ that combines the layout of a real floorplan with object and obstacle arrangements from scanned rooms.

\subsection{Partial Map Generation}
\label{sec:train_data_gene}
To generate training pairs from the composed scenes, we adopt the trajectory sampling approach from PONI. This method first generates shortest paths between random start and end positions, then samples observations along these paths. However, while PONI uses simple cropping to generate partial maps, we propose VisRC (Visibility through Ray Casting) to create partial observations subject to explicit visibility constraints.

\subsubsection{Trajectory Sampling}
Following PONI's approach, for each scene $G_{j,k}$, we sample exploration trajectories $\mathcal{T}_{j,k} = \{T_\tau\}_{\tau=1}^{|\mathcal{T}_{j,k}|}$ by selecting random start and end positions and computing shortest paths. Each trajectory $T_\tau = \{(x_t, y_t, o_t)\}_{t=1}^{|T_\tau|}$ represents a sequence of agent positions and orientations. For simplicity, we omit the scene indices $j,k$ when describing individual trajectories $T_\tau$ and coordinates $(x, y, o)$.

\subsubsection{VisRC Algorithm}
We introduce VisRC (Visibility through Ray Casting) to generate visibility-constrained partial maps that model field-of-view, sensing-range, and occlusion constraints during navigation. Unlike simple cropping methods that may reveal areas behind walls, VisRC simulates line-of-sight visibility based on the agent's pose and obstacle geometry. We set the field of view $\phi = 79^\circ$, maximum perception range $d_{\text{max}} = 5$m, and agent eye level $h_{\text{agent}} = 0.88$m, matching the agent configuration used during evaluation.

Let the agent be located at position $(x_t, y_t)$ with orientation $o_t$, field of view (FOV) $\phi$, and maximum perception range $d_{\text{max}}$. For each direction $\theta \in [-\phi/2, \phi/2]$, we define a ray as:
\begin{equation}
\text{Ray}(r, \theta) = (x_t, y_t) + r \cdot (\cos(o_t + \theta), \sin(o_t + \theta))
\end{equation}

This defines the trajectory along which the agent looks at angle $\theta$ from its current pose. The maximum visible distance in that direction is determined by the first point where the obstacle height exceeds the agent's eye level:
\begin{equation}
d(\theta) = \min\left(d_{\text{max}}, \min_r \left\{ r : O^{\text{full}}(\text{Ray}(r, \theta)) \geq h_{\text{agent}} \right\} \right)
\end{equation}

\begin{equation}
d_{\text{vis}}(\theta)=
\begin{cases}
d(\theta) & \text{if } \theta \in [-\phi/2, \phi/2] \\
0 & \text{otherwise}
\end{cases}
\end{equation}

Here, $O^{\text{full}}(x, y)$ denotes the obstacle height at location $(x, y)$ in the global map, represented as a 2.5D heightmap extracted from the 3D mesh (see the supplementary material for details). A point becomes occluded once its height reaches or exceeds the agent's eye level $h_{\text{agent}}$.

The visibility function $\text{Vis}(x, y; x_t, y_t, o_t)$ indicates whether a point $(x, y)$ is visible from the agent's current pose:
\begin{equation}
\text{Vis}(x, y; x_t, y_t, o_t) = 
\begin{cases}
1 & \text{if } \|(x, y) - (x_t, y_t)\| \leq d_{\text{vis}}(\theta) \\
0 & \text{otherwise}
\end{cases}
\end{equation}
where $\theta = \arctan2(y - y_t, x - x_t) - o_t$, $\text{Vis}(\cdot)$ defines the region that the agent can observe from its current pose $(x_t, y_t, o_t)$, accounting for occlusion and field-of-view limits. This function is used to construct visibility-constrained partial observations from global semantic and obstacle maps.

On 800 matched trajectories, comprising 10 trajectories in each of 80 HM3D scenes, the explored regions generated by VisRC achieve a mean IoU of 0.89 against regions accumulated from RGB-D observations under matched field-of-view, range, and occlusion conditions. Additional fidelity analyses are provided in the supplementary material.

\subsubsection{Training Pair Generation}

For each trajectory $T_\tau$, we simulate the agent's perception and build a sequence of partial maps $\{M_1^{T_\tau}, M_2^{T_\tau}, \ldots, M_{|T_\tau|}^{T_\tau}\}$ that reflect the agent's accumulated spatial knowledge over time.

At each timestep $t$, we compute the visible area using VisRC, which determines the set of points in the environment observable from the agent's current position and orientation based on ray casting. The explored map $E_t^{T_\tau}(x, y)$ at step $t$ is defined as the union of all visible regions up to that point:

\begin{equation}
E_t^{T_\tau}(x, y) = \bigvee_{t' = 1}^{t} \text{Vis}(x, y; x_{t'}, y_{t'}, o_{t'})
\end{equation}
where $\bigvee$ denotes the element-wise logical OR operator applied across all visibility maps from timestep 1 to $t$. Specifically, $E_t^{T_\tau}(x, y)$ is marked as explored if the position $(x, y)$ has been visible in at least one previous timestep along trajectory $T_\tau$.

We define the partial obstacle map $O_t^{T_\tau}(x, y)$ and partial semantic map $S_t^{T_\tau}(x, y)$ by masking the full maps with the explored region $E_t^{T_\tau}(x, y)$:
\begin{equation}
O_t^{T_\tau}(x, y) = E_t^{T_\tau}(x, y) \cdot O^{\text{full}}(x, y), \quad
\end{equation}

\begin{equation}
S_t^{T_\tau}(x, y) = E_t^{T_\tau}(x, y) \cdot S^{\text{full}}(x, y)
\end{equation}
where the multiplication is element-wise. This ensures that both obstacle and semantic information are included only for locations explored up to timestep $t$ along trajectory $T_\tau$.

Using these definitions, we construct the partial map $M_t^{T_\tau} \in \mathbb{R}^{H \times W \times (4 + N)}$ at timestep $t$ along trajectory $T_\tau$. The 4 channels encode partial obstacle map $O_t^{T_\tau}$, explored region $E_t^{T_\tau}$, trajectory history, and current position. The remaining $N$ channels represent the partial semantic map $S_t^{T_\tau}$. 

Each partial map $M_t^{T_\tau}$ is paired with the corresponding full semantic map $S^{\text{full}}_{k,j}$ of the scene to form a training pair, where $M_t^{T_\tau}$ serves as the input and $S^{\text{full}}_{k,j}$ acts as the ground truth. This pairing enables supervised learning of semantic map completion from partial observations collected along the agent's trajectory.
\section{Experiments}

\subsection{Experimental Setup}
\label{sec:exp_setup}

\subsubsection{Benchmarks and Datasets}
We evaluate NaviScale on two standard object navigation benchmarks: HM3D and MP3D. For HM3D, we use 80 training scenes and 20 validation scenes with 2000 validation episodes, targeting 6 object categories. For MP3D, we use 56 training scenes and 11 validation scenes with 2195 validation episodes, targeting 21 object categories. All validation scenes are held-out real 3D scans, providing a test of transfer from the composed training maps to unseen scanned environments. We reserve claims about physical deployment for the robot experiments in Section~\ref{sec:exp_real}.

\subsubsection{Baseline Methods}
We use PEANUT~\cite{zhai2023peanut} as our baseline for HM3D. For MP3D, we employ PEANUT combined with the Stubborn refine mapping module~\cite{luo2022stubborn}, as the lower 3D scanning quality in MP3D necessitates additional mapping refinement to ensure reliable navigation. Among the publicly available implementations we evaluated, these semantic map-based methods offer the most competitive performance on their respective datasets.

Our model architecture follows PEANUT's PSPNet design for the neural network $f$, which predicts target object locations from incomplete observed maps. To allow more training on the larger dataset, we use 300,000 iterations, a total batch size of 64 across four NVIDIA RTX 3090 GPUs, a 2,000-iteration warmup, and cosine learning-rate decay from $10^{-3}$ to $10^{-6}$. This replaces the shorter PEANUT-style schedule (60k iterations, extended to 150k in our initial experiments). These training settings are not a contribution of our method.
For MP3D, we further address class imbalance with Repeat Factor Sampling~\cite{gupta2019lvis} and a pixel-wise Seesaw loss~\cite{wang2021seesaw}, using $p=0.8$, $q=2.0$, and $\varepsilon=10^{-2}$; the supplementary material provides the complete formulation.
Unless otherwise specified, our results in simulation are averaged over five independent runs with different random seeds.

For the NaviScale framework configuration, except the ablation study Sec.~\ref{sec:exp_ab}, we set $K=8$ in our main experiments (Sec.~\ref{sec:exp_sota}, ~\ref{sec:exp_baseline} and ~\ref{sec:exp_data}), meaning each floorplan generates 8 full maps through composition, and each full map contributes one sampled trajectory to the training data.

\subsubsection{Evaluation Metrics}
We evaluate using two standard metrics: Success Rate (SR) and Success weighted by Path Length (SPL) which accounts for path efficiency by weighting success by the ratio of shortest path to actual path length.

\begin{figure*}[!t]
\centering
\includegraphics[width=2\columnwidth]{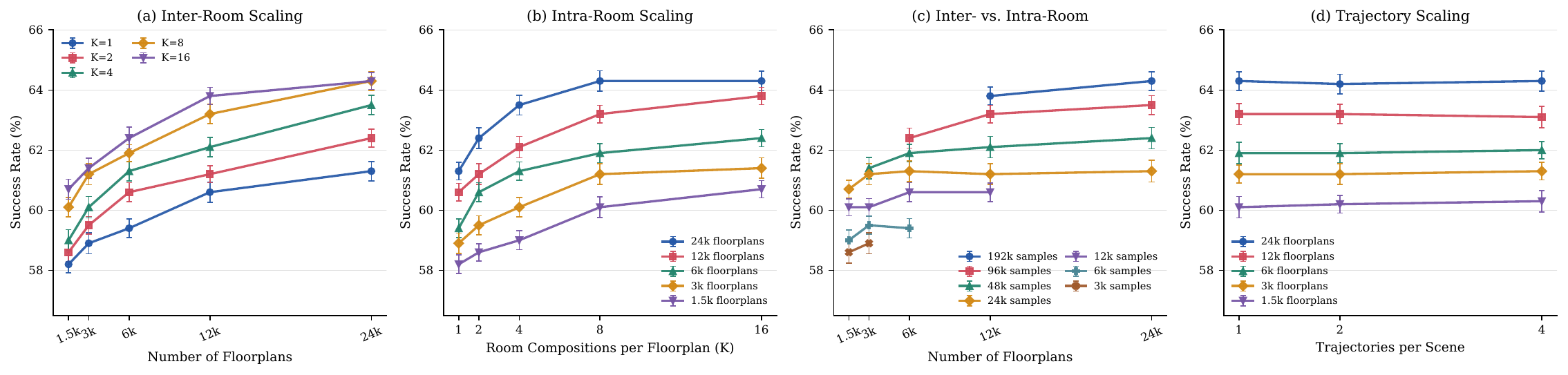}
\caption{Ablation study results for the scaling dimensions under the 300k training schedule. Each point reports the mean SR over five independent runs with different random seeds, and the error bars indicate one standard deviation.}
\label{fig:exp1}
\end{figure*}

\begin{table}[!t]
\caption{Object Navigation Results on HM3D and MP3D Benchmarks.
The upper block contains recent zero-shot systems, while the lower block contains task-trained map-based methods.
Results marked with * are reproduced using the authors' released code and weights.}
\label{tab:main_results}
\setlength{\tabcolsep}{2pt}
\centering
\makebox[\linewidth][c]{%
\begin{tabular}{l|c|cc|cc}
\hline
\multirow{2}{*}{Method} & \multirow{2}{*}{Venue} & \multicolumn{2}{c|}{MP3D} & \multicolumn{2}{c}{HM3D} \\
                         & & SR & SPL & SR & SPL \\
\hline
SG-Nav-GPT        & NeurIPS'24 & 40.2 & 16.0 & 54.0 & 24.9 \\
BeliefMapNav          & NeurIPS'25 & 37.3 & 17.6 & 61.4 & 30.6 \\
CoS                   & AAAI'26    & 37.6 & 17.6 & 55.9 & 29.1 \\
TrajRAG               & CVPR'26    & 42.6 & 18.0 & 62.5 & 33.9 \\
GLMap                 & CVPR'26    & 42.5 & 18.3 & 62.7 & 33.7 \\
\hline
SGM                   & CVPR'24    & 37.7 & 14.7 & 60.2 & 30.8 \\
PEANUT                & ICCV'23    & -    & -    & 58.8* & 29.1* \\
G3D-LF                & CVPR'25    & 39.0 & 18.8 & 55.6 & 31.8 \\
\hline
Ours (NaviScale, 300k) &           & 43.1 & 16.8 & 64.3 & 34.8 \\
\hline
\end{tabular}%
}
\end{table}

\begin{table}[!t]
  \caption{Effects of Training Data and Training and Inference Settings. Results are means over five independent runs. Within each benchmark, the rows marked Updated use the same architecture and training and inference settings and differ only in training data.}
  \label{tab:data_protocol_attribution}
  \setlength{\tabcolsep}{3pt}
  \centering
\makebox[\linewidth][c]{%
\begin{tabular}{l|l|l|cc}
  \toprule
  Benchmark & Training data & Protocol & SR$\uparrow$ & SPL$\uparrow$ \\
  \midrule
  HM3D & HM3D & Updated & 59.6 & 29.8 \\
  HM3D & NaviScale & Initial & 61.5 & 31.0 \\
  HM3D & NaviScale & Updated & \textbf{64.3} & \textbf{34.8} \\
  \midrule
  MP3D & MP3D & Updated & 40.3 & 14.1 \\
  MP3D & NaviScale & Initial & 40.3 & 14.0 \\
  MP3D & NaviScale & Updated & \textbf{43.1} & \textbf{16.8} \\
  \bottomrule
  \end{tabular}%
}
\end{table}

\begin{table}[!t]
  \caption{Dataset Comparison on HM3D Validation. Results are means over five independent runs with different random seeds. PT: pretrain, FT: finetune.}
  \label{tab:dataset_comparison}
  \centering
\makebox[\linewidth][c]{%
\begin{tabular}{l|cc}
  \hline
  Training Data & SR & SPL \\
  \hline
  ProcTHOR & 55.2 & 24.0 \\
  MP3D only & 58.6 & 29.2 \\
  HM3D only & 59.6 & 29.8 \\
  MP3D + HM3D & 61.9 & 33.7 \\
  \hline
  Ours  & 64.3 & 34.8 \\
  Ours + MP3D + HM3D & 64.2 & 34.8 \\
  Ours PT \& MP3D+HM3D FT & 64.3 & 35.5 \\
  \hline
  \end{tabular}%
}
  \end{table}

\subsection{Comparison with State-of-the-Art Methods}
\label{sec:exp_sota}
We compare our method with two categories of approaches: (1) recent zero-shot systems, including SG-Nav~\cite{yin2024sgnav}, BeliefMapNav~\cite{zhou2025beliefmapnav}, Chain-of-Search (CoS)~\cite{chen2026chain}, TrajRAG~\cite{wang2026trajrag}, and GLMap~\cite{zhang2026glmap}; and (2) task-trained map-based methods, including SGM~\cite{zhang2024imagine}, PEANUT~\cite{zhai2023peanut}, and G3D-LF~\cite{wang2025g3d}.

Table~\ref{tab:main_results} compares navigation systems on MP3D and HM3D; our system retains the same prediction architecture. Under the updated 300k schedule, HM3D reaches 64.3\% SR and 34.8\% SPL, exceeding the strongest listed prior SR (GLMap, 62.7\%) by 1.6 percentage points. On MP3D, our system reaches the highest listed SR of 43.1\%, while G3D-LF reports the highest listed SPL of 18.8\%.

The system-level comparison alone does not separate the contribution of training data from changes to training and inference. Table~\ref{tab:data_protocol_attribution} therefore holds the architecture and these settings fixed within each benchmark: replacing the original training data with NaviScale improves HM3D by 4.7 SR and 5.0 SPL points and MP3D by 2.8 SR and 2.7 SPL points. Comparing the Initial and Updated rows for NaviScale instead shows the gain from longer training and the other training and inference changes; implementation details are provided in the supplementary material.

Notably, our approach outperforms several recent LLM-based methods without invoking an LLM at inference time or changing the underlying navigation architecture, suggesting that scaling domain-specific spatial data can be an effective alternative to introducing general-purpose language priors for navigation.

\subsection{Comparison with Other Datasets}
\label{sec:exp_data}
Having isolated the gain from replacing the original training data, we next compare NaviScale with other data sources and their combinations. For MP3D and HM3D datasets, we follow PEANUT's data processing pipeline, which generates training data by having an agent navigate in 3D environments and build maps from observations. Our VisRC method approximates this process on 2D semantic maps through ray-casting simulation. For ProcTHOR, we first extract top-down semantic maps from the simulator, then apply our training data generation method described in Section~\ref{sec:train_data_gene} to create the training set. Table~\ref{tab:dataset_comparison} reports the five-run mean for each training-data configuration on the HM3D validation set using the same network architecture.

ProcTHOR remains lowest at 55.2\% SR. Among the real datasets, MP3D and HM3D reach 58.6\% and 59.6\% SR, respectively, while combining them reaches 61.9\% SR. NaviScale achieves the best performance when each dataset is used alone for training at 64.3\% SR. Compared with the rule-driven procedural dataset ProcTHOR, NaviScale improves the five-run mean SR by 9.1 points (55.2\% to 64.3\%) and SPL by 10.8 points (24.0\% to 34.8\%) under the same prediction architecture. This result supports using floorplans and room layouts from real environments to generate navigation training data. Adding real datasets to NaviScale reaches 64.2\% SR, providing no additional gain and suggesting that the compositional dataset already captures the spatial regularities needed by the predictor.

\subsection{Effect on Different Baselines}
\label{sec:exp_baseline}

\begin{table}[!t]
\caption{Performance of Two Navigation Baselines on HM3D Validation When Trained with HM3D or NaviScale. Results are means over five independent runs with different random seeds.}
\label{tab:different_baseline}
\centering
\makebox[\linewidth][c]{%
\begin{tabular}{l|l|cc}
\hline
Method & Training Data & SR & SPL \\
\hline
L3MVN & HM3D & 53.4 & 24.1 \\
L3MVN & Ours & 55.1 & 25.8 \\
\hline
PEANUT & HM3D & 59.6 & 29.8 \\
PEANUT & Ours & 64.3 & 34.8 \\
\hline
\end{tabular}%
}
\end{table}

The preceding comparisons vary the training data for a fixed model. To examine whether the benefit extends to another architecture, we train two object navigation models, L3MVN and PEANUT, using either the original HM3D training set or our generated dataset, and evaluate them on the HM3D validation split. For each model--data combination, Table~\ref{tab:different_baseline} reports the mean over five independent runs. NaviScale improves performance across both architectures without requiring model changes.

\subsection{Ablation Studies}
\label{sec:exp_ab}
We next examine which choices in data generation contribute to these gains. Table~\ref{tab:data_quality} first tests how room categories and room-to-mask fitting affect the composed maps under the 300k training schedule. We then vary the number of floorplans and room compositions to study data scaling, before comparing how partial observations are generated.

\textbf{Room-category assignment.} Manual annotation assigns functional categories to floorplan room masks. The LLM baseline instead classifies the room descriptions extracted from the SVG files, while random assignment replaces these category assignments with random labels. These labels determine which source room maps are eligible for composition. Manual annotation exceeds LLM classification by 4.9 SR points and random assignment by 20.1 points. The comparison supports the importance of reliable category matching: changing a room's assigned category changes the semantic and obstacle configurations used to populate it.

\textbf{Room-to-mask fitting.} Size-constrained fitting selects only source room maps smaller than the target mask. Crop-to-fit permits larger source maps but removes portions outside the mask. Scale-to-fit rescales selected room maps to fit the target, subject to the scale-ratio filtering described in the supplementary material. Scale-to-fit achieves 64.3\% SR, compared with 63.6\% for size-constrained fitting and 59.8\% for cropping. A plausible explanation is that scaling retains complete local configurations, whereas cropping can remove objects and their spatial context; size restrictions also reduce the eligible source maps. The experiment compares these fitting strategies as complete procedures.

\begin{table}[!t]
\caption{Data-Generation Quality Ablations Under the 300k Training Schedule.}
\label{tab:data_quality}
\centering
\makebox[\linewidth][c]{%
\begin{tabular}{lcc}
\toprule
Configuration & SR & SPL \\
\midrule
Manual annotation & 64.3 & 34.8 \\
LLM classification & 59.4 & 30.8 \\
Random assignment & 44.2 & 24.0 \\
\midrule
Scale-to-fit & 64.3 & 34.8 \\
Size-constrained & 63.6 & 34.0 \\
Crop-to-fit & 59.8 & 30.4 \\
\bottomrule
\end{tabular}%
}
\end{table}

\subsubsection{Inter-Room Scaling}

Figure~\ref{fig:exp1}(a) demonstrates the impact of inter-room scaling under different intra-room scaling levels. Each curve represents a fixed $K$ value (room composition number) while varying the number of floorplans from 1.5k to 24k. The results show consistent performance improvements across all $K$ values as floorplan numbers increase, indicating the effectiveness of inter-room scaling.

\subsubsection{Intra-Room Scaling}
Figure~\ref{fig:exp1}(b) examines intra-room scaling effects by fixing floorplan numbers and varying $K$ from $1$ to $16$. The results show consistent performance improvements as room composition diversity increases across all floorplan scales. Notably, performance gains plateau at higher $K$ values, with $K=8$ to $K=16$ at 24k floorplans showing no further improvement (both achieving 64.3\% SR), suggesting saturation in room composition diversity benefits.

\subsubsection{Inter-Room vs. Intra-Room Scaling}

To compare the two scaling dimensions, we constrain the number of generated maps and vary its allocation in Figure~\ref{fig:exp1}(c). Each curve fixes $K \times \text{floorplans}$, while its points trade more distinct floorplans against more recompositions of each floorplan. The generally upward trend suggests that, under these budgets, using more distinct floorplans is more effective than increasing $K$. The gap grows at larger budgets. This comparison evaluates floorplan diversity as a whole, including differences in room count, size, shape, connectivity, corridors, and overall layout.

\begin{figure*}[!t]
\centering
\includegraphics[width=2\columnwidth]{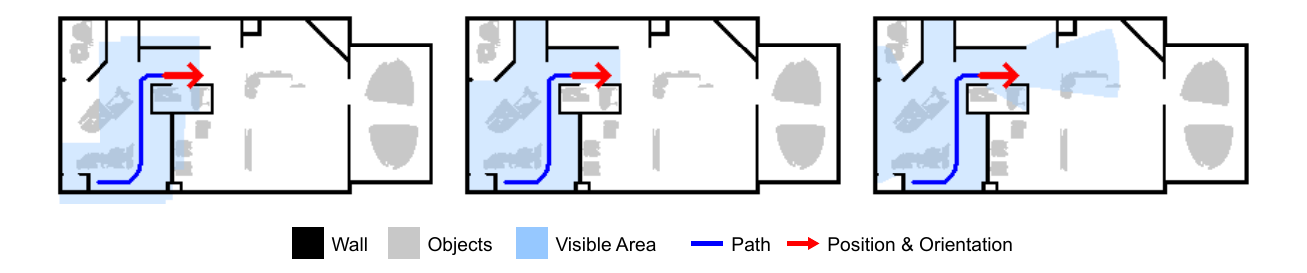}
\caption{Comparison of partial map generation methods. Left: crop along path (PONI); Middle: room visible along path; Right: our VisRC method.}
\label{fig:visrc}
\end{figure*}

\subsubsection{Trajectory Scaling}

We investigate trajectory scaling by varying the number of
trajectories per scene with fixed $K=8$. As shown in
Figure~\ref{fig:exp1}(d), the performance curve remains relatively   
flat, indicating limited benefits. This saturation effect can be attributed to the fact that each trajectory already generates multiple training samples at every step within the same scene. Since all samples from a single scene share the same spatial layout and semantic structure, adding more trajectories in the same scene provides redundant information without improving the model's ability to generalize to new environments.

\begin{table}[!t]  
\caption{Comparison of Partial-Map Generation Methods on HM3D Validation Under the 300k Training Schedule. All methods share the same underlying trajectories. Random and path-based crops match VisRC's final cumulative visible area within 5\% per trajectory; room-visible observations cover approximately 12\% more area overall.}
\label{tab:partial_map_methods}
\centering
\makebox[\linewidth][c]{%
\begin{tabular}{l|cc}
\hline
Method & SR & SPL \\
\hline
Random crop & 60.2 & 28.3 \\
Crop along path & 60.8 & 33.4 \\
Room visible along path & 62.6 & 34.2 \\
VisRC (Ours) & 64.3 & 34.8 \\
\hline
\end{tabular}%
}
\end{table}

\subsubsection{Partial Map Generation Methods}

We compare VisRC against three baseline methods using the same underlying trajectories: Random crop places rectangular observations at randomly selected map locations; Crop along path follows PONI's cropping strategy with an additional area-matching procedure; and Room visible along path reveals complete rooms when entered. For each trajectory, we first generate the VisRC observations as the reference. For each crop baseline, we fix the crop centers and relative crop sizes, then iteratively adjust a common scale factor, enlarging or shrinking the crops until their cumulative visible area at the end of the trajectory differs from VisRC's by at most 5\%. Cumulative area is measured over the union of all revealed regions, counting overlapping cells only once. This constraint matches final coverage.

The room-visible method preserves whole-room observations and is not area-matched; its cumulative visible area is approximately 12\% larger than VisRC's overall. As shown in Table~\ref{tab:partial_map_methods}, VisRC performs best, including against the two crop baselines with matched final coverage. This supports the value of visibility-constrained partial-map generation beyond simply increasing the final observed area.

Room visible along path outperforms the area-matched Crop along path baseline by 1.8 percentage points in SR. Path-based crops provide fragmented observations, whereas the room-visible strategy presents complete observations of entered rooms while leaving other rooms unobserved. This result is consistent with the value of unseen-room prediction, although the larger observed area in the room-visible condition can also contribute to the difference. The fixed-map-budget allocation study provides separate evidence for the benefit of floorplan-level structural diversity.

\subsection{Analysis and Discussion}
Our scaling studies show that increasing distinct floorplans yields greater gains than increasing room recomposition diversity ($K$) when the total number of generated maps is held fixed in our experiments. Separately, a partial-map method that exposes complete rooms and leaves others unobserved outperforms one based on fragmented intra-room crops. Together, these experiments provide empirical evidence that floorplan-level structural diversity and supervision for unseen-room prediction are valuable in semantic map-based ObjectNav.

The distinction can be understood through two ways of acquiring spatial information: direct observation through exploration and inference from learned spatial regularities. Within the current room, turning the camera or moving a few steps can often reveal much of the layout and its objects. For example, an agent searching for a television in a living room may inspect the area around a visible sofa with a short local scan. Some furniture remains occluded, but much of the information supplied by intra-room prediction may soon become available through relatively inexpensive observation.

Other rooms remain hidden behind walls even when the agent turns in place. Observing them requires finding a doorway, moving through it, and potentially following a corridor. If the target is absent, the agent may then need to backtrack and inspect another room. Predicting which unseen region is likely to contain the target can guide this choice before the travel cost is incurred. For example, kitchen-related evidence may help prioritize an adjacent region when searching for a dining-area object. The potential benefit of inter-room prediction is therefore its ability to direct exploration across rooms under a limited navigation budget.

This difference in exploration cost offers an explanation for the fixed-budget result: additional floorplans expose the predictor to more arrangements of rooms that cannot be inspected through a short local scan. Intra-room information remains useful, but some of it can be acquired directly as the agent explores its current room. The room-visible comparison is consistent with this interpretation, although its larger observed area may also contribute to the difference. Thus, the experiments support prioritizing floorplan diversity in our setting, while the exploration-cost argument explains why this allocation may help.

\subsection{Physical-Robot Transfer Demonstration}
\label{sec:exp_real}

\begin{figure}[!t]
\centering
\includegraphics[width=\linewidth]{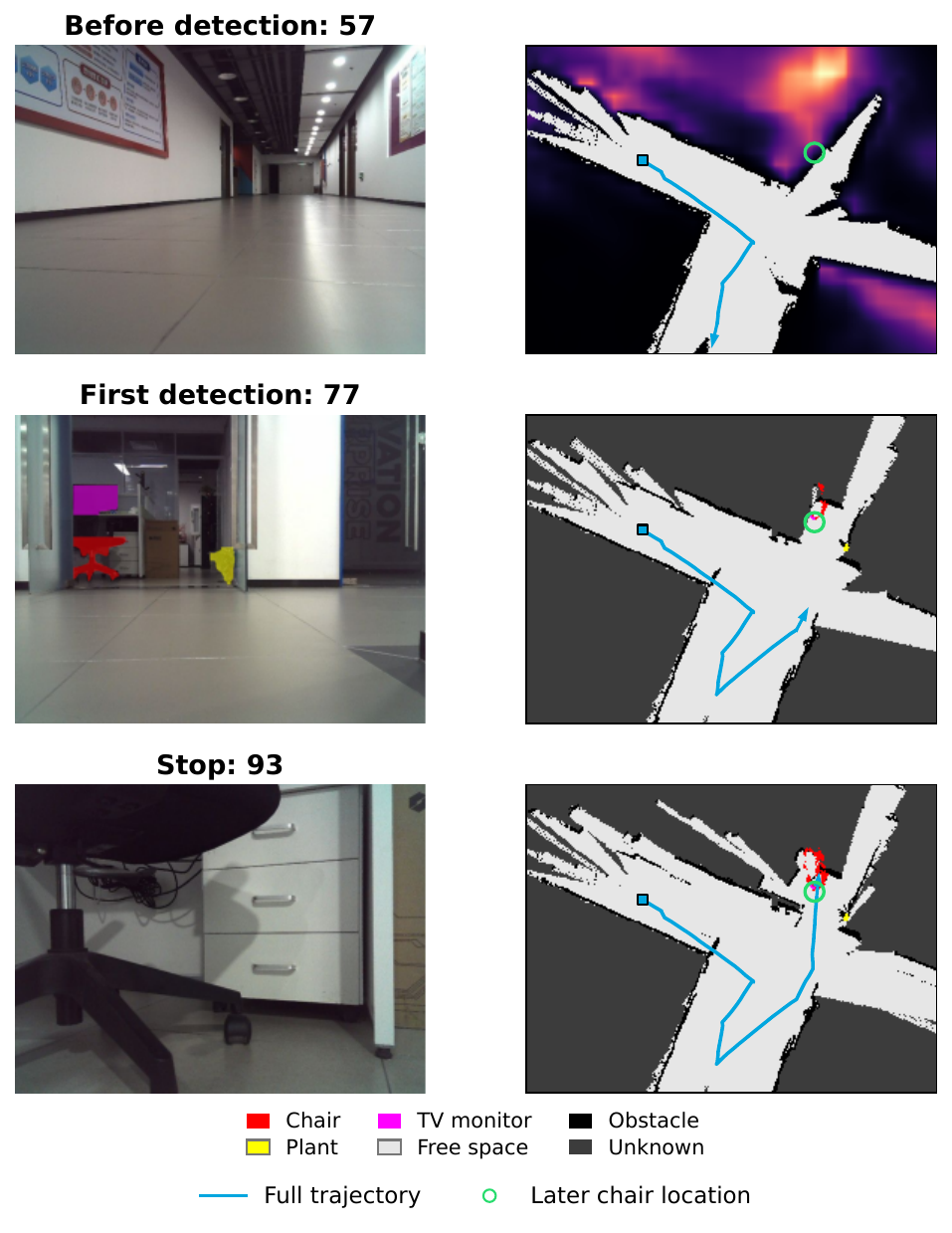}
\caption{Physical chair-search sequence at recorded states 57, 77, and 93. Left: RGB images with recorded segmentation overlays (chair in red). Right: accumulated semantic maps over the same map area with full recorded trajectories in blue. The first row overlays unfiltered chair prediction scores on unknown cells: brighter yellow/orange indicates higher scores, while darker purple/black indicates lower scores. The second RGB view shows part of an office area containing two rows of desks and chairs. The highest-response region also contains a chair, distinct from the one subsequently selected for navigation. Green circles mark the chair location obtained from later semantic observations, not surveyed ground truth. The last two rows show target detection and navigation using the previously mapped chair location without prediction overlays. In the final frame, only part of the chair is visible and no chair mask is produced, but its previously mapped location remains available.}
\label{fig:real_chair_episode}
\end{figure}

The preceding experiments evaluate navigation in scanned environments. We now examine whether the trained pipeline can also operate on a physical robot with online perception and mapping. We use a Hiwonder JetAuto wheeled robot in two indoor scenes. S1 is an approximately 50~m$^2$ home evaluated over 12 episodes, while S2 is an approximately 200~m$^2$ office evaluated over 20 episodes collected in two batches of 10. In every run, the semantic map is constructed online from RGB-D observations using RTAB-Map rather than supplied as ground truth. All computation is performed on one RTX 3090 GPU. We report success rate and FPS. The FPS values are not directly comparable to simulation because each physical action cycle takes approximately one second due to robot execution and sensor-update latency.

\begin{table}[!t]
  \caption{Physical-Robot Transfer Demonstration with a Hiwonder JetAuto.}
  \label{tab:realworld}
  \centering
\makebox[\linewidth][c]{%
\begin{tabular}{llcc}
    \toprule
    Scene & Method & Succ$\uparrow$ & FPS$\uparrow$ \\
    \midrule
    S1 (12 ep.) & PEANUT~\cite{zhai2023peanut} & 58.3 & 1.2 \\
                & PEANUT + NaviScale            & 75.0 & 1.2 \\
    S2 (20 ep.) & PEANUT                         & 40.0 & 1.2 \\
                & PEANUT + NaviScale            & 60.0 & 1.2 \\
    \bottomrule
  \end{tabular}%
}
\end{table}

Table~\ref{tab:realworld} shows gains in both environments: NaviScale improves success by 16.7 percentage points in S1 and 20 percentage points in S2. No visually evident mapping drift or distortion was observed during these trials. Given 32 episodes across two sites, these results establish deployment feasibility; broader robustness evaluation requires additional environments and robot platforms.

Figure~\ref{fig:real_chair_episode} illustrates a separately recorded chair-search episode on a fixed $960\times960$ map grid at 5-cm resolution. At recorded state 57, before any chair detection, the highest prediction response points toward a region containing another chair in the two-row office arrangement, rather than the chair ultimately selected for navigation. At state 77, the robot detects a chair and starts navigating toward its location projected onto the map. At state 93, the robot stops after reaching the stopping-distance threshold for the previously mapped chair location, despite the absence of a new chair mask in the current frame.

\subsection{Limitations and Future Directions}
Increasing $K$ produces new combinations of source room maps but does not expand the set of available source room maps. Rare arrangements absent from the MP3D and HM3D source library therefore remain underrepresented even as the number of composed maps grows. Extending the source library is a complementary way to increase intra-room diversity. The current dataset also emphasizes residential floorplans and static scene configurations; broader building types and dynamic interactions require additional sources and evaluation.

\section{Conclusion}

We presented NaviScale, a framework for generating large-scale semantic map training data that addresses data scarcity in semantic map-based ObjectNav by expanding the training data from hundreds of scenes to 192,000 synthetic semantic maps constructed from real sources. The framework combines floorplan-level inter-room structural diversity with intra-room semantic and obstacle recomposition, while VisRC converts full maps into visibility-constrained partial observations. With 300k training iterations and the training and inference settings described in this paper, the system reaches 64.3\% SR and 34.8\% SPL on HM3D, together with 43.1\% SR and 16.8\% SPL on MP3D, without modifying the prediction architecture.
When the total number of generated maps is fixed, using more distinct floorplans yields larger gains than repeated recompositions or trajectories within the same layouts. These findings highlight floorplan-level structural diversity and unseen-room prediction as promising directions for expanding navigation training data.

\bibliographystyle{IEEEtran}
\bibliography{NaviScale}

\clearpage
\title{Supplementary Material for ``NaviScale: Generating Large-Scale Semantic Map Datasets for Object Navigation''}
\author{Chuanlin Lan, Yanwei Zheng, Weijian Liu, Zhitong Zhou, Xiao Zhang,\\
Fuzhen Zhuang, and Dongxiao Yu}
\maketitle

\section{Data Construction Details}

\subsection{Floorplan Collection and Ethics}

We collected 12,794 unique residential properties from publicly accessible real-estate listings. Multi-story properties were split by floor, producing 24,000 floorplans. Room geometry and room descriptions are used only during dataset construction. The public release contains only de-identified training data: rasterized obstacle maps, sampled trajectories, and object-category annotations introduced by our composition process. It excludes addresses, listing URLs, photographs, property identifiers, source floorplan files, natural-language room descriptions, and room-category annotations; source-to-output links are also removed. Because obstacle maps retain coarse spatial geometry, we describe this release as de-identified rather than claiming irreversible anonymization. This data-minimization strategy reduces re-identification risk while preserving the information required for navigation training. Room-mask segmentation and labeling required about 1,300 human-hours. The dataset primarily represents North-American residential properties and therefore may not cover other geographic regions, architectural styles, commercial buildings, or uncommon room configurations.

\subsection{Unified Room Categories}

After consolidating Transition regions into Others, the floorplans use 12 labels. Eleven labels specify the room categories used when composing room maps; Others is an auxiliary label for regions that do not receive a source room map. Table~\ref{tab:supp_room_mapping} combines the MP3D native-label mapping with the category distribution of rooms in the MP3D and HM3D ObjectNav v1 training splits. MP3D provides native room labels. HM3DSem v0.1 provides semantic regions but no native room-category labels, so we manually assign room-category labels to its valid semantic regions using the unified 12-label vocabulary. These labels are human annotations rather than automatically inferred categories. HM3DSem v0.2 is used only for the dataset-scale and annotation-cost comparison in the main paper.

\begin{table*}[!t]
\caption{Unified Room-Category Mapping and Training-Split Room Distribution. MP3D counts use native room labels, whereas HM3D counts use manually assigned room-category labels because HM3DSem v0.1 does not provide native room-category labels. The two training splits contain 56 MP3D scenes and 80 HM3D scenes, respectively.}
\label{tab:supp_room_mapping}
\centering
\makebox[\linewidth][c]{%
\begin{tabular}{l>{\raggedright\arraybackslash}p{0.49\textwidth}rrr}
\toprule
Merged category & MP3D native labels & MP3D & HM3D & Total \\
\midrule
Bedroom & bedroom & 159 & 271 & 430 \\
Living & living room, tv, lounge, familyroom/lounge & 113 & 97 & 210 \\
Dining & dining room, dining booth, bar & 45 & 18 & 63 \\
Kitchen & kitchen & 49 & 89 & 138 \\
Bathroom & bathroom, toilet, spa/sauna & 256 & 257 & 513 \\
Work\&Study & office, library, classroom, meeting/conference room & 94 & 74 & 168 \\
Recreation & rec/game, workout/gym/exercise & 16 & 10 & 26 \\
Utility & laundry/mud room, utility/tool room, junk, closet & 103 & 49 & 152 \\
Garage & garage & 8 & 8 & 16 \\
Stairs\&Hall & stairs, hallway & 311 & 84 & 395 \\
Outdoors & outdoor, balcony, porch/terrace/deck, entry/foyer/lobby & 156 & 6 & 162 \\
Others & other room & 54 & 223 & 277 \\
\midrule
Total & -- & 1,364 & 1,186 & 2,550 \\
\bottomrule
\end{tabular}%
}
\end{table*}

Table~\ref{tab:supp_room_distribution} reports the number of base room masks in each functional category before generating the $K$ room-map compositions for every floorplan. The category counts sum to 301,663 room masks across 24,000 floorplans.

\begin{table}[!t]
\caption{Room-Category Distribution Across the 24,000-Floorplan Dataset.}
\label{tab:supp_room_distribution}
\centering
\makebox[\linewidth][c]{%
\begin{tabular}{lr}
\toprule
Category & Base room masks \\
\midrule
Bedroom & 47,478 \\
Living & 26,479 \\
Dining & 14,518 \\
Kitchen & 13,896 \\
Bathroom & 48,319 \\
Work\&Study & 4,300 \\
Recreation & 2,476 \\
Utility & 98,833 \\
Garage & 7,390 \\
Stairs\&Hall & 20,254 \\
Outdoors & 17,720 \\
\midrule
Total & 301,663 \\
\bottomrule
\end{tabular}%
}
\end{table}

\subsection{Room Maps, Alignment, and Connectivity}

We project semantic labels and geometry from the MP3D and HM3DSem v0.1 training splits onto a top-down grid with 5-cm cells. Each semantic cell stores an object category. The obstacle representation is a 2.5D heightmap in which each cell stores the maximum obstacle height. This representation supports both visibility calculations at the height of the agent's camera and traversability checks at the agent body height.

For a floorplan room mask, we compare the mask's maximum inscribed rectangle with the source room map's minimum bounding rectangle. Source maps that fit are placed without excessive deformation; moderately oversized maps are scaled to fit; maps requiring a scale ratio above 1.5 are rejected and resampled. After composition, connected components are computed over navigable cells. A composition is invalid when its rooms are not connected through navigable space; the affected room assignments are then resampled.

The 24,000 unique floorplans cover 3,381,405~m$^2$. Because each floorplan produces eight composed maps at $K=8$, the aggregate footprint counted across the 192,000 generated maps is 27,051,240~m$^2$.

\section{Evaluation of Generated Maps and Sim-to-Real Transfer}

\subsection{Evaluation Scope}

A semantic map has two roles: its semantic labels are used to predict target locations, and its obstacle channels support FMM planning. Geometry errors can affect both prediction and navigation, whereas semantic-label errors directly affect prediction and target recognition. Table~\ref{tab:supp_map_roles} distinguishes the three map sources evaluated in this work.

These map sources raise two practical questions: whether the generated training maps provide useful spatial supervision, and whether the trained system can operate with maps built from sensor observations. For the first question, we describe the composition constraints and compare VisRC with RGB-D visibility along the same HM3D trajectories; the main paper evaluates the resulting training data through navigation performance. For the second, we compare ground-truth and segmented semantic maps below, while the main paper reports physical deployment with online RGB-D mapping.

\begin{table}[!t]
\caption{Geometry Source, Semantic Source, and Use of Each Map Representation.}
\label{tab:supp_map_roles}
\setlength{\tabcolsep}{3pt}
\centering
\makebox[\linewidth][c]{%
\begin{tabular}{llll}
\toprule
Map & Geometry & Semantics & Use \\
\midrule
NaviScale & Composed & Assigned labels & Training \\
HM3D/MP3D & Scanned & Segmented & Pred. + nav. \\
Real world & Measured & Segmented & Pred. + nav. \\
\bottomrule
\end{tabular}%
}
\end{table}

Here, ``Assigned labels'' are the semantic labels defined by each composed training map; they are not ground truth for a corresponding physical environment.

\subsection{Quality of Generated Maps}

\subsubsection{Recomposition and Traversability}

NaviScale combines two sources of spatial information from real environments: floorplans supply inter-room topology and room adjacency, while room maps supply object co-occurrence and obstacle geometry within rooms. Matching room categories prevents placing the contents of one room type into another, scale-ratio filtering limits geometric deformation, and connectivity validation rejects blocked layouts. The representation follows the agent model with a fixed radius and height used by semantic-map navigation benchmarks. We therefore evaluate whether the generated maps are useful for navigation, rather than whether they reconstruct photorealistic scenes or exactly reproduce the distribution of real buildings and their contents.

\subsubsection{Agreement with RGB-D Observations}

On matched HM3D trajectories, we compare the explored region generated by VisRC with the region accumulated from RGB-D observations. The mean explored-region IoU is 0.89 over 800 trajectories, comprising 10 trajectories in each of 80 HM3D scenes. This experiment evaluates the visibility component of the generated partial maps under matched field-of-view, range, and occlusion conditions.

\subsection{Sim-to-Real Transfer Evaluation}

\subsubsection{Semantic-Segmentation Errors}

To distinguish errors in training supervision from errors encountered during navigation, we vary the semantic-label source separately for NaviScale training maps and HM3D test maps. ``GT'' uses ground-truth labels. ``Seg.'' maps are accumulated by traversing a scene, segmenting RGB observations, fusing them into a semantic map, and extracting room maps when needed. In Table~\ref{tab:supp_semantic_noise}, replacing GT with segmented training maps changes navigation performance little under the same test condition. Replacing GT test labels with segmented labels produces a larger success-rate drop, while prediction distances remain close. This pattern is consistent with target misrecognition being a greater difficulty than predicting unseen target locations.

\begin{table}[!t]
\caption{Effects of Ground-Truth and Predicted Semantic Labels on HM3D Under the 300k Training Schedule.}
\label{tab:supp_semantic_noise}
\centering
\makebox[\linewidth][c]{%
\begin{tabular}{llccc}
\toprule
Train & Test & $d_{\mathrm{pred}}$ (m)$\downarrow$ & SR$\uparrow$ & SPL$\uparrow$ \\
\midrule
GT & GT & 0.89 & 70.9 & 38.9 \\
GT & Seg. & 0.92 & 64.3 & 34.8 \\
Seg. & GT & 0.92 & 70.7 & 38.8 \\
Seg. & Seg. & 0.91 & 64.2 & 35.0 \\
\bottomrule
\end{tabular}%
}
\end{table}

\section{Qualitative Examples}

\begin{figure}[!t]
\centering
\includegraphics[width=\linewidth]{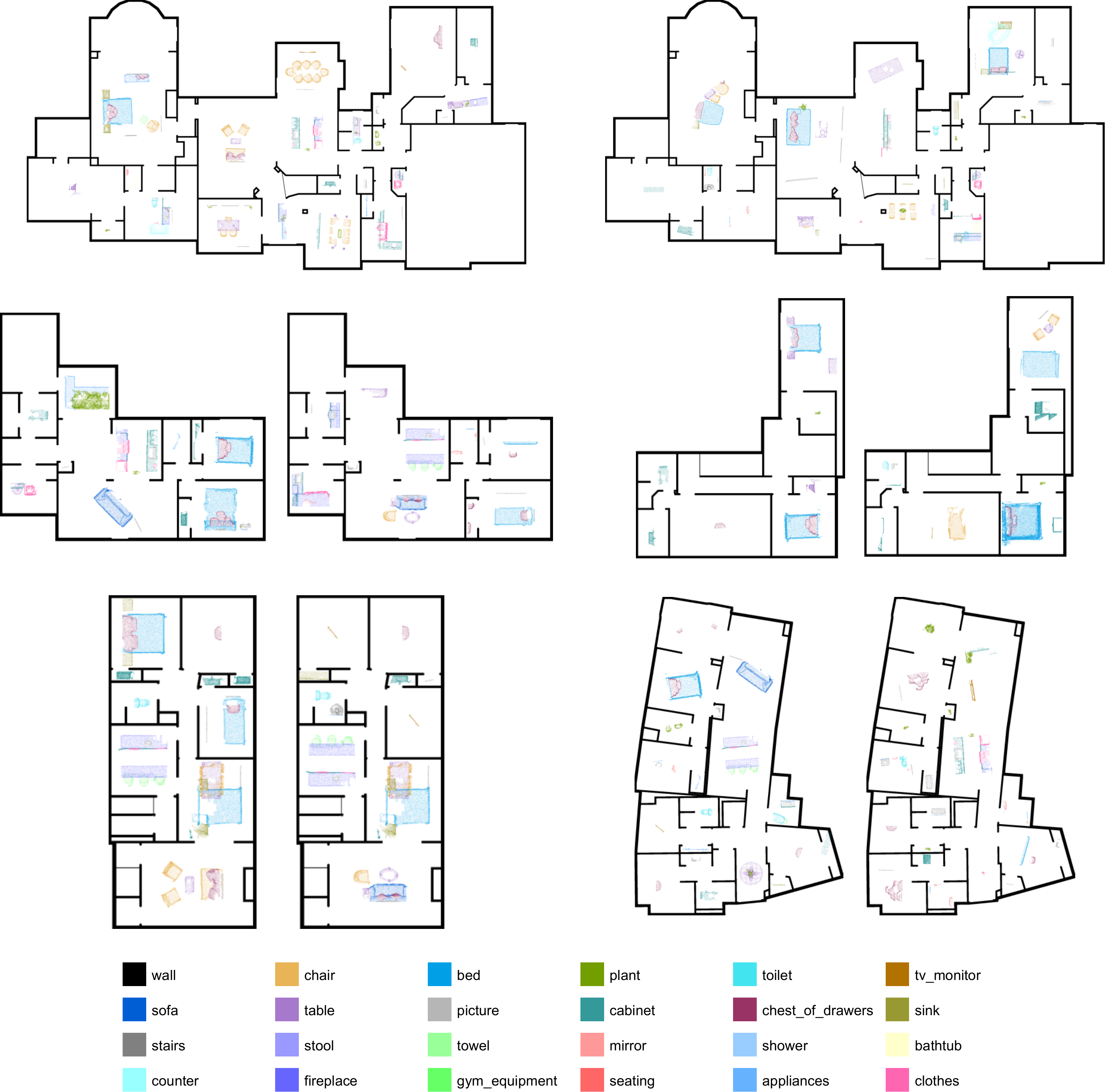}
\caption{Examples of complete semantic maps generated by composing room maps with matching room categories within floorplan layouts.}
\label{fig:supp_semantic_examples}
\end{figure}

\begin{figure}[!t]
\centering
\includegraphics[width=\linewidth]{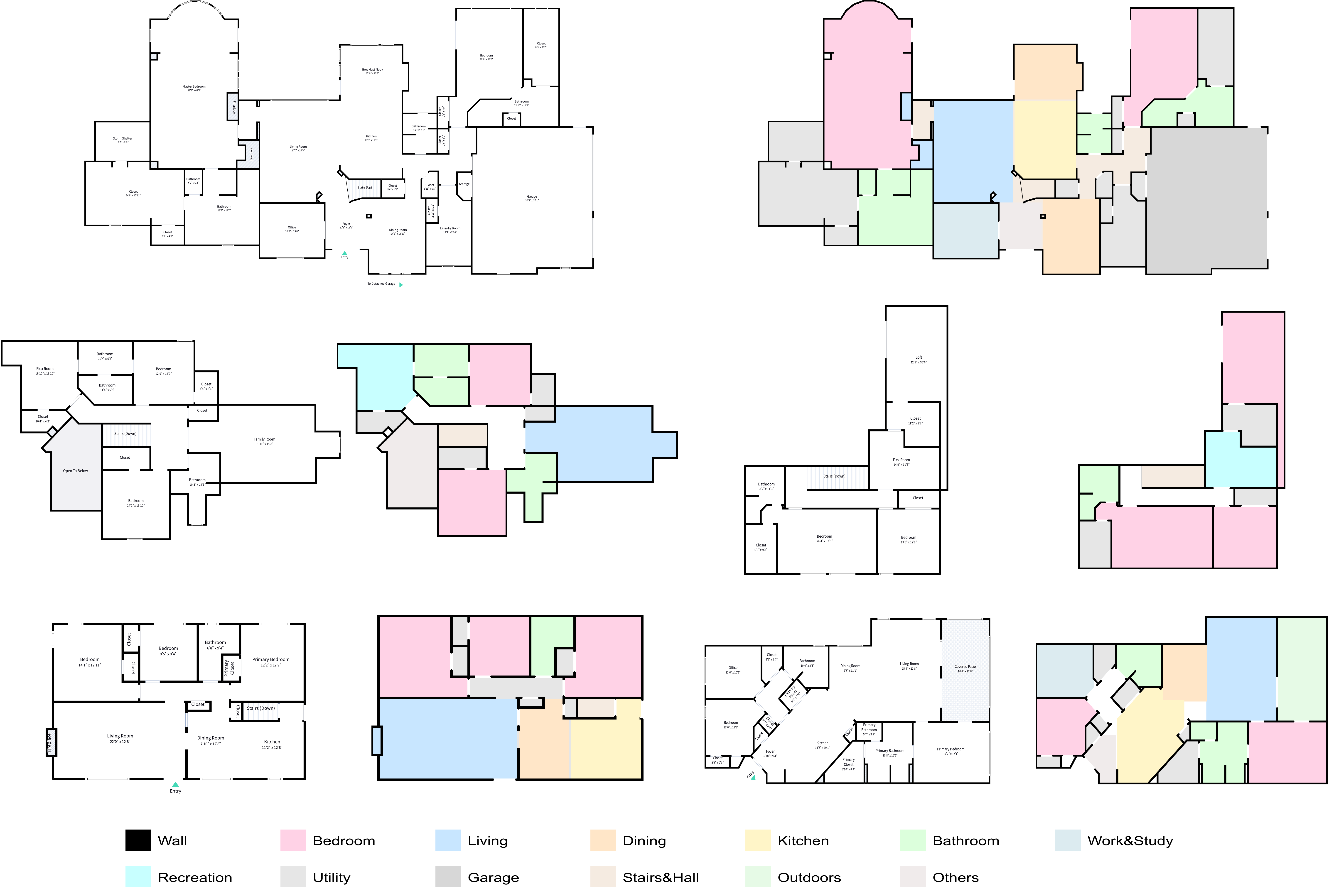}
\caption{Examples of floorplans with room-category labels under the unified 12-label vocabulary.}
\label{fig:supp_room_examples}
\end{figure}

\begin{figure}[!t]
\centering
\includegraphics[width=\linewidth]{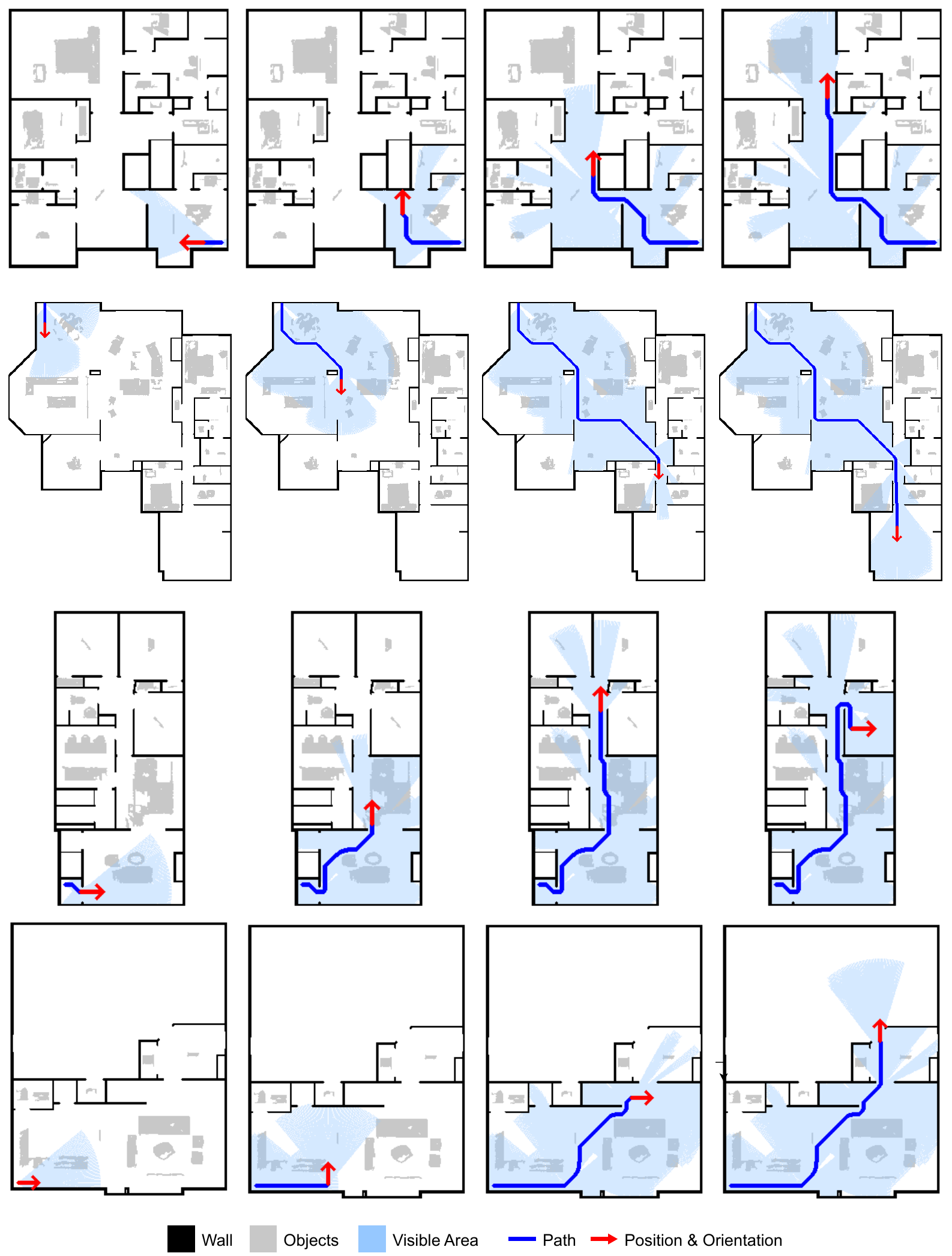}
\caption{Examples of visibility-constrained partial observations accumulated along VisRC trajectories.}
\label{fig:supp_trajectory_examples}
\end{figure}

Figures~\ref{fig:supp_semantic_examples}--\ref{fig:supp_trajectory_examples} show generated complete semantic maps, floorplans with room labels, and partial observations accumulated along VisRC trajectories, respectively.

\section{Training and Inference Protocol Details}

\subsubsection{Segmentation Models and Training Images}

For HM3D, we follow the segmentation component released with PEANUT: a Mask R-CNN pretrained on COCO and adapted using images sampled exclusively from HM3D training scenes. For MP3D, we follow the RedNet segmentation setup used by PONI and further train the segmenter using RGB observations rendered from MP3D training scenes and their corresponding semantic labels. In both benchmarks, all scene-sampled images used for segmentation training or adaptation come exclusively from the respective training split; no images or labels from validation or test scenes are used for this training. Reconstructed MP3D scenes exhibit rendering artifacts, including fragmented geometry, missing surfaces, and texture discontinuities. These introduce an appearance gap relative to generic pretraining images, motivating adaptation to the rendered observations rather than direct use of the pretrained checkpoint. The class-balancing procedure is detailed below. NaviScale's composed semantic maps train the navigation predictor; they are not RGB training images for the segmenter.

For the separately recorded physical-robot episode shown in the main paper, the robot uses YOLOv8s-seg with the checkpoint file \texttt{yolov8s-seg.pt} and a segmentation confidence threshold of 0.25. The segmenter used in this robot episode differs from those used in the simulation experiments. The confidence thresholds and required numbers of supporting frames described next apply to MP3D.

\subsubsection{Multi-Frame Target Confirmation}

At each step, the semantic segmentation output is projected into a current-frame semantic map and compared with the accumulated historical map. A frame supports an existing object candidate when their same-class regions overlap in map coordinates. Supporting frames need not be consecutive; their evidence is accumulated over time for the same mapped object.

Detections below 0.90 confidence are ignored directly. Confidence in $[0.90,0.95)$ requires at least three supporting frames, confidence in $[0.95,0.98)$ requires at least two, and confidence at or above 0.98 requires only the current frame. Before the required count is reached, the candidate region is masked from target selection and the agent searches elsewhere, while its historical evidence remains available for subsequent accumulation. Once the count reaches the threshold, the candidate is selected and the agent navigates toward it.

\subsubsection{Class-Balanced Segmentation Training}

Repeat Factor Sampling increases the frequency of rare target categories. For category $c$, its image-level frequency is
\begin{equation}
f_c=\frac{\left|\{I:c\in Y_I\}\right|}{|\mathcal{I}|},
\end{equation}
where $\mathcal{I}$ is the training-image set and $Y_I$ is the set of categories present in image $I$. With repeat threshold $t=0.001$, the category-level and image-level repeat factors are
\begin{equation}
r_c=\max\left(1,\sqrt{\frac{t}{f_c}}\right),
\end{equation}
\begin{equation}
r_I=\max_{c\in Y_I}r_c.
\end{equation}
Thus, an image containing multiple categories is repeated according to the largest repeat factor among its categories.

We additionally apply Seesaw loss to every valid semantic-segmentation pixel. For pixel $n$ with ground-truth class $i=y_n$, the loss is
\begin{equation}
\ell_n=-\log
\frac{\exp(z_{n,i})}
{\exp(z_{n,i})+\sum_{j\ne i}M_{ij}C_{n,ij}\exp(z_{n,j})},
\end{equation}
where $z_{n,j}$ is the logit for class $j$. The class-frequency mitigation factor is
\begin{equation}
M_{ij}=\begin{cases}
\left(N_j/N_i\right)^p, & N_j<N_i,\\
1, & N_j\geq N_i,
\end{cases}
\end{equation}
where $N_i$ and $N_j$ are the training-set pixel counts of classes $i$ and $j$. The prediction-dependent compensation factor is
\begin{equation}
C_{n,ij}=\begin{cases}
\left(\dfrac{\pi_{n,j}}{\max(\pi_{n,i},\varepsilon)}\right)^q,
& \pi_{n,j}>\pi_{n,i},\\
1, & \pi_{n,j}\leq\pi_{n,i},
\end{cases}
\end{equation}
with $\pi_{n,j}=\exp(z_{n,j})/\sum_{k=1}^{C}\exp(z_{n,k})$. We use $p=0.8$, $q=2.0$, and $\varepsilon=10^{-2}$. The final loss averages all non-ignored pixels $\mathcal{V}$:
\begin{equation}
L_{\mathrm{Seesaw}}=\frac{1}{|\mathcal{V}|}\sum_{n\in\mathcal{V}}\ell_n.
\end{equation}
The mitigation factor protects rare classes from excessive suppression by frequent classes, while the compensation factor increases the penalty when an incorrect class is currently more probable than the ground-truth class.

Together with a 2,000-iteration warmup, cosine learning-rate decay, and 300k training iterations, these class-balancing and target-confirmation components yield the navigation result of 43.1 SR and 16.8 SPL on MP3D. The main paper compares training datasets using the same architecture and training and inference settings.

\end{document}